\documentclass[10pt,twocolumn,letterpaper]{article}

\usepackage{iccv}
\usepackage{times}
\usepackage{epsfig}
\usepackage{graphicx}
\usepackage{amsmath}
\usepackage{amssymb}

\usepackage[pagebackref=true,breaklinks=true,letterpaper=true,colorlinks,bookmarks=false]{hyperref}

\usepackage{caption}
\usepackage{adjustbox}
\usepackage{wrapfig}
\usepackage{multirow}
\usepackage{makecell}

\usepackage{algorithm}
\usepackage{algpseudocode}
\usepackage{kotex}
\usepackage[numbers,sort]{natbib}
\usepackage{verbatim}

\usepackage{booktabs}

\iccvfinalcopy 

\def\iccvPaperID{1935} 

\ificcvfinal\fi

\begin{document}

\title{Toward Spatially Unbiased Generative Models}
\author{Jooyoung Choi$^1$ ~~~~~~~ Jungbeom Lee$^1$ ~~~~~~~ Yonghyun Jeong$^3$  ~~~~~~~  Sungroh Yoon$^{1,2, }$\thanks{Correspondence to: Sungroh Yoon (sryoon@snu.ac.kr)}\\
$^1$ Data Science and AI Laboratory, Seoul National University, Korea\\
$^2$ ASRI, INMC, and Interdisciplinary Program in AI, Seoul National University, Korea\\
$^3$ AI Research Team, Samsung SDS\\
{\tt\small \{jy\_choi, jbeom.lee93, sryoon\}@snu.ac.kr ~~~~~~~ yhyun.jeong@samsung.com}}
\ificcvfinal\thispagestyle{empty}\fi

\maketitle


\begin{abstract}
   Recent image generation models show remarkable generation performance. However, they mirror strong location preference in datasets, which we call \textbf{spatial bias}. Therefore, generators render poor samples at unseen locations and scales. We argue that the generators rely on their implicit positional encoding to render spatial content. From our observations, the generator’s implicit positional encoding is translation-variant, making the generator spatially biased. To address this issue, we propose injecting explicit positional encoding at each scale of the generator. By learning the spatially unbiased generator, we facilitate the robust use of generators in multiple tasks, such as GAN inversion, multi-scale generation, generation of arbitrary sizes and aspect ratios. Furthermore, we show that our method can also be applied to denoising diffusion probabilistic models. Our code is available at: \url{https://github.com/jychoi118/toward_spatial_unbiased}.
\end{abstract}

\thispagestyle{empty}
\section{Introduction}

Recent CNN-based generative models~\cite{karras2017progressive, stylegan,brock2018large,ho2020denoising,karras2020analyzing} generate images of remarkable quality by learning the distribution of well-designed datasets. 
The careful design of a dataset~\cite{yu2015lsun, stylegan}, along with architecture design~\cite{karras2017progressive,stylegan,brock2018large}, is an important factor~\cite{esser2020note} in the development of generative model performance.
First, humans carefully collect high-quality images~\cite{stylegan}. Then, to benefit learning, the collected data are pre-processed according to human inductive bias. Pre-processing may contain center-cropping~\cite{yu2015lsun}, resizing, mirror padding, and alignment based on landmark annotation~\cite{karras2017progressive,kazemi2014one}. Eventually, the images may have a strong location preference, which we call \textbf{\textit{spatial bias}}, due to the photographer's bias in the initial data collection and pre-processing of the collected data. Both are used so the model learns the underlying concepts of the dataset.

\begin{figure}[t!]
    \centering
    \includegraphics[width=1.0\linewidth]{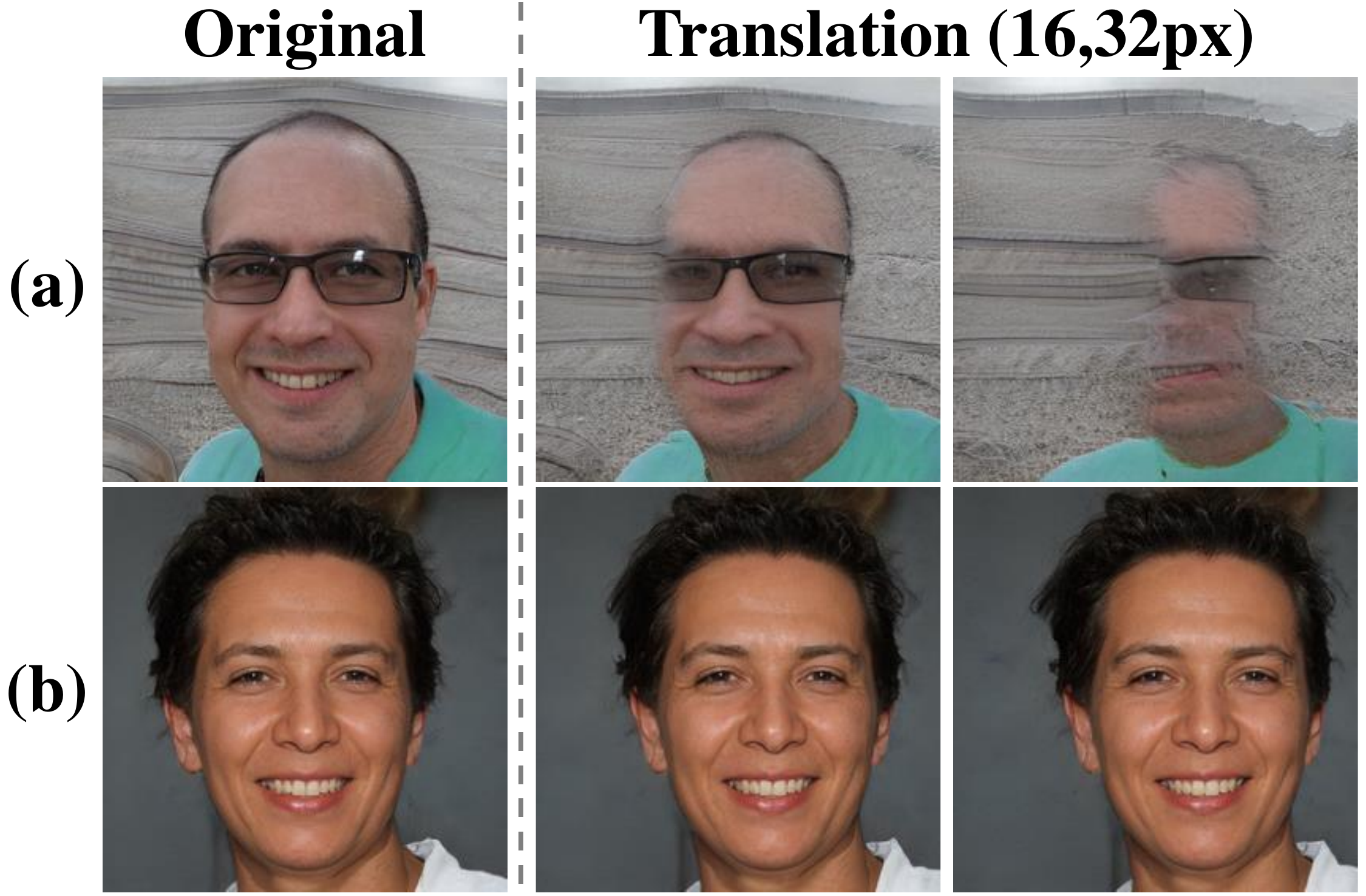}
    \caption{\textbf{Generation at translated locations.} (a) Baseline StyleGAN2; (b) Spatially unbiased StyleGAN2. $256^2$ images generated at 16, 32 pixels rolled locations. (a) shows distorted and harmed samples. With our method, a model can generate well-structured images in unseen locations.}
    \label{fig:teaser}
    \vspace{-1em}
\end{figure}

The collection and processing of the dataset are intended to improve learning, but the resulting spatial bias can be an unintended shortcut for the model. If the spatial bias is a cue that helps the model learn the data distribution easily, the model can rely on this shortcut~\cite{bahng2020learning} despite the sufficient model capacity. 
We demonstrate that generator components, such as constant tensors and zero-padding, jointly work as implicit positional encoding. Since they are located at fixed absolute positions, such implicit positional encoding is translation-variant. Thus, it encourages the model to rely on spatial bias.
This phenomenon may be undesirable from a generalization perspective. 
As seen in Fig.~\ref{fig:teaser}(a), a spatially biased model generates distorted and harmed samples at shifted positions.

Therefore, our goal is to utilize a well-designed dataset to learn rich representations without relying on spatial bias.
We explore implicit positional encoding of the model, analogous to studies in discriminative tasks~\cite{islam2020much,kayhan2020translation}. At each scale of the generative models, positional information introduced from the previous scale and zero-padding in the current scale jointly work as implicit encoding. To render spatial content, the generator relies on implicit positional encoding. We demonstrate that this implicit cue is translation-variant and therefore suffers from spatial bias. To avoid reliance on implicit cues, we propose injecting an explicit scale-specific positional embedding into the generator. Our method learns spatially unbiased and translation-equivariant generators, which generate content according to a given positional encoding.

Since we provide explicit scale-specific positional information, the generator no longer needs to rely on its implicit cue. Therefore, the generator creates content according to the given positional information, which enables translation-equivariant generation.
Fig.~\ref{fig:teaser}(b) shows consistent facial features generated at unseen positions. We illustrate that multi-scale positional encoding enables the generation of an image at an unseen position by manipulating the positional embedding at the inference time.

Concurrent to our work, Karras~\textit{et al.}~\cite{karras2021alias}
observe that previous GANs are not translation-equivariant and the details stick to the absolute location. Based on signal processing, \cite{karras2021alias} introduce multiple architectural changes to enable the coarsest feature to control the position of finer features. In constrast, we control positions with explicit scale-specific embeddings, without further architectural changes.

Through extensive experiments, we demonstrate that multi-scale positional encoding is effective in learning spatially unbiased generators. With only minimal modification in model architectures, our method maintains the generation quality of previous well-performing models, as shown by comparison of Fréchet Inception Distance (FID)~\cite{heusel2017gans} as well as Precision and Recall~\cite{sajjadi2018assessing}. Nevertheless, our method leads to the spatially unbiased generator. Removing spatial bias further encourages extension to various applications. We demonstrate robustness to affine transformations in GAN inversion~\cite{zhu2016generative,brock2016neural,yeh2017semantic,abdal2019image2stylegan,gu2020image} and image edition~\cite{shen2021closedform,abdal2019image2stylegan,gu2020image}. We show the effectiveness in multi-scale generation with a model trained on a single resolution. Our method enables the generation of arbitrary sizes and aspect ratios by expanding and extrapolating positional embedding in the test time. Furthermore, we also apply our method on Denoising Diffusion Probabilistic Models (DDPM)~\cite{ho2020denoising,sohl2015deep}, which is a recent family of generative models. The main contributions of this work are threefold:

\begin{itemize}

\item We discuss that existing generators rely on translation-variant implicit positional encoding, making the generators spatially biased.

\item We propose a method for learning a spatially unbiased generator with a minimal modification in well-performing architectures.

\item We validate our method on multiple tasks:  GAN inversion, multi-scale generation, generation of arbitrary sizes and aspect ratios, and adaptation to DDPM.

\end{itemize}

\section{Related Work}

\begin{figure*}[t!]
    \centering
    \includegraphics[width=1.0\linewidth]{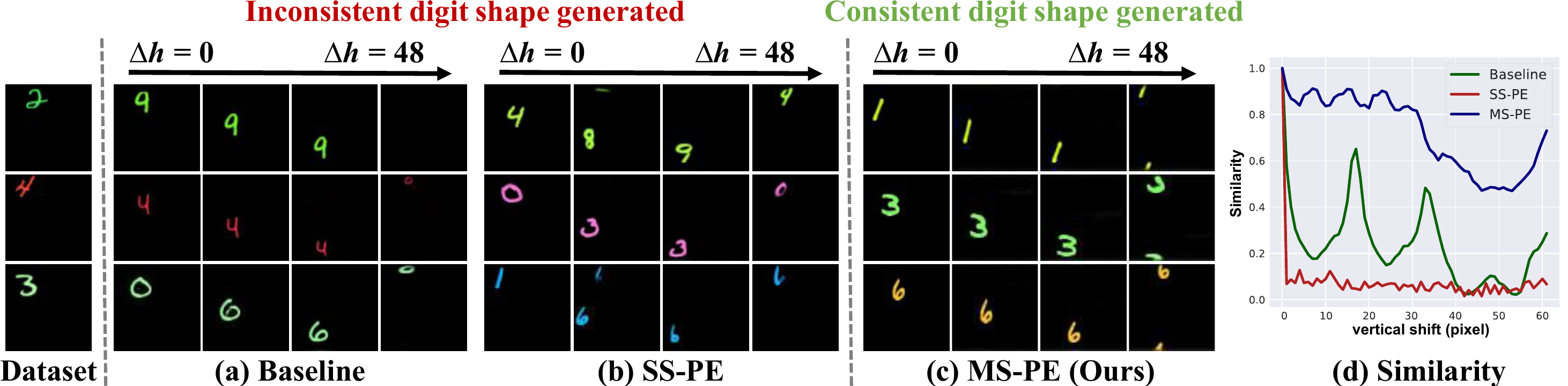}
    \caption{\textbf{Generation at shifted locations.} We generated digits at locations shifted vertically by $\triangle h$, sharing the same inputs. We trained generators on spatially biased images shown on the left. (a) and (b) exhibit inconsistent digits at shifted locations, while our method (c) generates consistent digits. (d) exhibits the similarity measured during successive translations.}
    \label{fig:toy}
\end{figure*}

\subsection{Translation Variance of CNN}

Translation invariance is an essential feature of convolution operations with advantages in computer vision tasks such as robustness of classification~\cite{zhang2019making} and object detection~\cite{manfredi2020shift}. However, despite the translation-invariant convolutional operation, modern CNN architectures are not translation-invariant. Various reasons have been studied: photographer's bias~\cite{manfredi2020shift} in a dataset, zero-padding implicitly providing absolute positional information~\cite{xu2020positional,islam2020much,kayhan2020translation}, and lack of consideration of the sampling theory in downsampling operations~\cite{azulay2018deep,zhang2019making}. The resulting translation-variant model can cause inconsistent inferences depending on the input position~\cite{zhang2019making,manfredi2020shift,azulay2018deep,alsallakh2020mind,kayhan2020translation}.

\subsection{Positional Encoding}

Positional encoding plays a simple yet powerful role in various domains such as natural language processing (NLP) and computer vision. In the transformer architecture~\cite{vaswani2017attention}, 
the order of the sequence is provided by adding sinusoidal embedding to the input feature. Similarly, the transformer-based object detection model DETR~\cite{carion2020end} informs the spatial order of features at each location with 2D sinusoidal embedding.
Several studies~\cite{carion2020end,dosovitskiy2020image} have shown that the presence of positional embedding significantly improve the performance, while different embedding variants (e.g., sinusoidal embedding, Cartesian grid, learned constant) yield similar performances. Positional encoding also plays an important role in 3D vision studies. NeRF~\cite{mildenhall2020nerf} maps input coordinates to a high-dimensional space with sinusoidal embedding to learn high-frequency 3D representations.

\subsection{Bias in Generative Models}

Recent studies~\cite{menon2020pulse,esser2020note,zhao2018bias} on the bias of generative models examine that the model can reflect the bias of the learned dataset. Esser~\etal\cite{esser2020note} demonstrated that a high-quality dataset leads to high-quality generation by disentangling data-biased and unbiased content. 
In PULSE~\cite{menon2020pulse}, a super-resolution study with a StyleGAN~\cite{stylegan} model, it was discovered that there is a facial color bias in the generated images due to severe imbalance in the dataset and GAN's mode-collapse behavior. 
In contrast, we consider spatial bias. Huh \etal~\cite{huh2020transforming} found an object-center bias in the BigGAN~\cite{brock2018large}. Therefore, it was difficult to project a real image into a latent space. The paper addressed this using a separate transformation optimization method. Analogous to our work, Xu \etal~\cite{xu2020positional} found that zero-padding works as implicit positional encoding in SinGAN~\cite{shaham2019singan}, then provided explicit encoding at the coarsest scale of the model. In contrast, we investigate implicit encoding at each scale and observe that explicit encoding at the coarsest scale is insufficient for learning spatially unbiased models.

\section{Method}

Before we analyze the spatial bias of generative models, we introduce the background of StyleGAN~\cite{stylegan,karras2020analyzing}, which is our baseline, in Sec.~\ref{sec:3.1}. In Sec.~\ref{sec:3.2}, we introduce multi-scale positional encoding (MS-PE), the method of learning spatially unbiased generative model. Sec.~\ref{sec:3.3} shows that existing generators rely on implicit positional encoding, which is translation-variant, and that our method addresses this issue. We describe how to generate at an unseen position and resolution in Sec.~\ref{sec:3.4}.

\begin{figure*}[t!]
    \centering
    \includegraphics[width=1.0\linewidth]{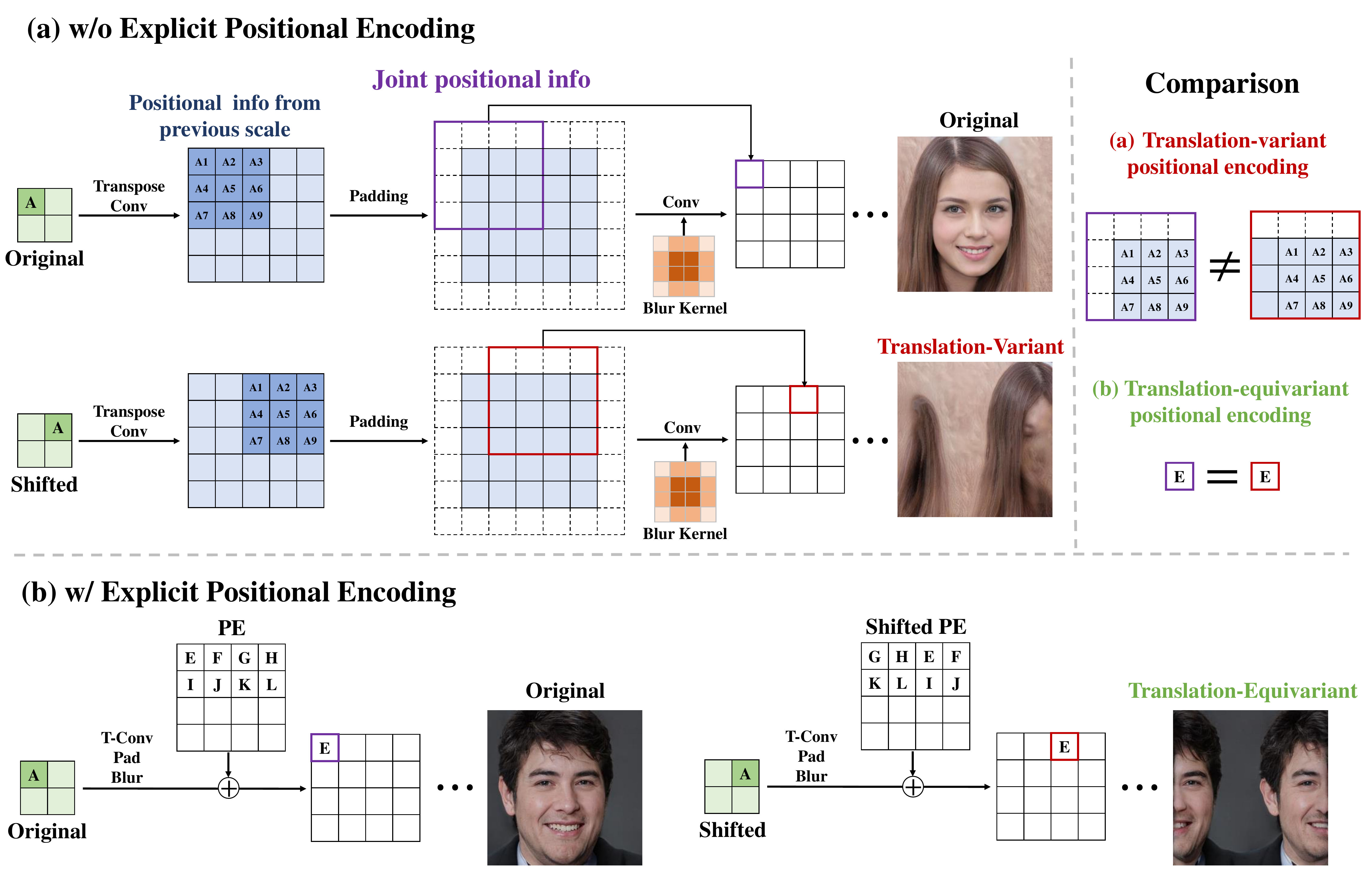}
    \caption{\textbf{Illustrative example of shifted generation.} We illustrate how implicit and explicit positional encoding affect generation at shifted location. Alphabets represent location indices. \textbf{(a)} Positional information is extracted from previous scale and zero-padding. Shift in previous scale causes inconsistent shift of positional encoding in current scale. Therefore destructive image is generated. \textbf{(b)} Positional information is given explicitly. We simply shift explicit encoding according to the shift of the previous scale, thus provide consistent positional encoding. }
    \label{fig:single-pe}
\end{figure*}

\subsection{Preliminary}\label{sec:3.1}

To analyze the spatial bias of the CNN-based generation models, we focus on StyleGAN~\cite{stylegan,karras2020analyzing}, where three inputs are provided: 1D latent vector $w$, 2D noise map $\epsilon$, and a learnable constant tensor $c$. Each is known to serve as globally coherent style~\cite{stylegan,shen2020interpreting},  spatial details~\cite{stylegan}, and implicit positional encoding~\cite{xu2020positional} respectively.

First, we investigate the spatial bias of StyleGAN using our customized dataset. We create a spatially biased color-MNIST, where digits are located only in the upper-left corner. 
In our experiments regarding shifts, we used a circular shift where elements shifted beyond the boundary are ``rolled'' to the other side.

The original StyleGAN, which we denote as ``baseline'', generates inconsistent shapes from rolled constant tensor $c$, as shown in Fig.~\ref{fig:toy}(a). Thus, we can infer that a constant tensor $c$ does not serve as an accurate positional encoding.

\subsection{Multi-Scale Positional Encoding}\label{sec:3.2}
In this section, we introduce a method of providing scale-specific positional information, which we call \textit{multi-scale positional encoding} (MS-PE).
Let the $L$-scale be a generator with scale $l\in \{ 1,...,L\}$, spatial dimensions $H^{l}, W^{l}$, channel number $C^{l}$ and feature $h^{l}$. 
We use a simple yet powerful sinusoidal positional embedding widely used in transformer literatures~\cite{vaswani2017attention,carion2020end,dosovitskiy2020image}. Sinusoidal positional embedding is a predefined code consisting of different sine or cosine function values for each location. We add the code to the convolutional feature. The dimension is equal to the number of feature channels. The code representing the location $(i,j)$ is as follows:
\begin{equation}\label{eq:posenc}
PE_{(i,j)}=[pe_{(i,0)},...,pe_{(i,2d-1)},pe_{(j,0)},...,pe_{(j,2d-1)}],
\end{equation}
\begin{equation}\label{eq:pe}
pe_{i,2k}=\sin(i/10000^{k/2d}), ~pe_{i,2k+1}=\cos(i/10000^{k/2d}),
\end{equation}
where $k=0,1,...,d-1$ and $d$ is a quarter of the channel dimension. $PE^{l}$ is the positional embedding of the $l$-th scale, a set of $PE_{(0,0)},...,PE_{(H^{l},W^{l})}$. Note that $PE^{l}$ and $h^{l}$ have the same dimensionality.

Additionally, we multiply the learnable scalar $\gamma^l$ before adding the positional encoding of each scale to the feature. Learnable scalars allow the generator to adjust the scale of positional information as desired. The features at each scale are calculated as follows:
\begin{equation}\label{eq:scalar}
h^{l+1}=f^{l}(PE^{l},w^{l},h^l)=G^{l}(w^{l},h^l)+\gamma ^{l} PE^{l},
\end{equation}
where $G^{l}$ and $w^{l}$ are the convolutional layer and latent vector of the $l$-th scale, respectively. Note that the feature at each scale is a function of the latent vectors and positional embeddings. As an exception, positional encoding is given as a feature input in the coarsest scale:
\begin{equation}\label{eq:replace}
h^{1}=G^{0}(PE^{0},w^{0}).
\end{equation}
Xu~\etal~\cite{xu2020positional} also provided explicit positional encoding as an input tensor in the coarsest scale to offer spatial inductive bias to a generator. However, in the next section, we demonstrate that this is insufficient, and that the multi-scale positional encoding (MS-PE) of Eq.~\ref{eq:scalar} is required.

Combining Eq.~\ref{eq:scalar} and Eq.~\ref{eq:replace}, the generative process can be denoted as:
\begin{equation}\label{eq:process}
I=f^{L}(PE^{L},w^{L},f^{L-1}(...,f^{0}(PE^{0},w^{0}))).
\end{equation}

\subsection{Why Does MS-PE Work Better?}\label{sec:3.3}

Before we demonstrate the effectiveness of MS-PE, we first study the implicit positional encoding of the generator.
Fig.~\ref{fig:single-pe} exhibits an example sequence of transpose convolution and anti-aliasing (blur)~\cite{zhang2019making} of StyleGAN~\cite{stylegan,karras2020analyzing} upsampling a $2\times 2$ feature to a $4\times 4$ feature.
From location A in $2\times 2$ feature, upsampling sequence introduces detailed positional information A1--9. Then, the shape of zero-paddings~\cite{islam2020much} is encoded by anti-aliasing convolution. Positional information A1--9 from the previous scale and zero-padding jointly serve as implicit positional encoding, providing unique positional information to each location. These implicit cues are located in fixed absolute locations. 
As a consequence, implicit positional encoding at shifted location differs from that at the original location, as illustrated in the top-right corner of Fig.~\ref{fig:single-pe}. The different shape of zero-padding is encoded in the shifted location. Therefore, the implicit positional encoding is translation-variant. 

Baseline StyleGAN's learnable constant tensor $c$ and zero-padding are known to jointly work as implicit positional encoding ~\cite{xu2020positional}. 
The generator renders spatial content relying on these implicit cues.
However, since the generator's implicit cue is translation-variant 
, it leads to poor generation quality at shifted locations, as shown in Fig.~\ref{fig:single-pe}(a).  
In other words, the generator is spatially biased, failing to generalize to unseen positions. The generator renders reasonable images only at the learned positions.
Fig.~\ref{fig:teaser}(a) and Fig.~\ref{fig:toy}(a) show distorted samples at unseen positions.

\begin{figure}[t!]
    \centering
    \includegraphics[width=0.9\linewidth]{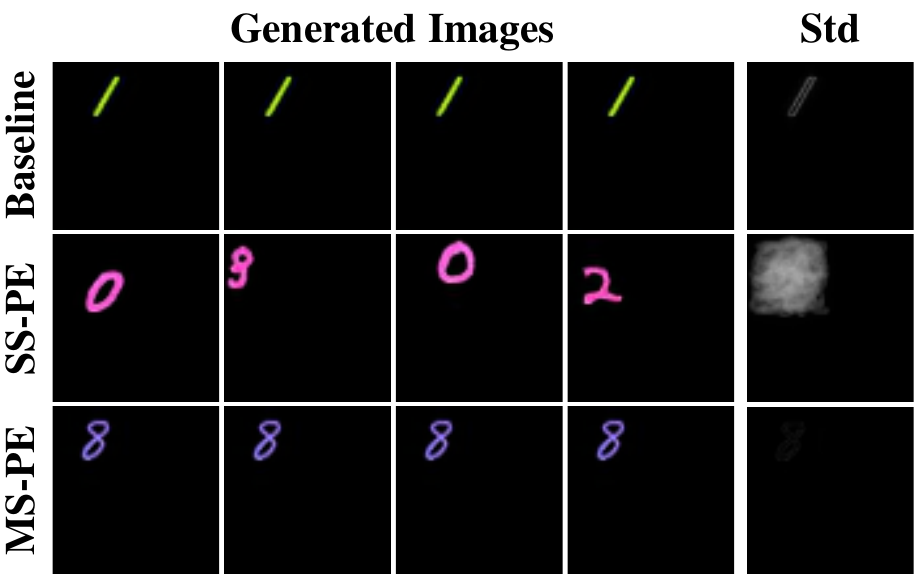}
    \caption{\textbf{Effect of 2D noise input.} Standard deviation of each pixel is calculated over 100 2D noise map instances. SS-PE generate inconsistent shapes and the largest standard deviations compared to the baseline and MS-PE.}
    \label{fig:toy-noise}
\end{figure}

Our motivation is to provide positional information explicitly to prevent the generator from relying on implicit cues. As a starting point, we replace the constant tensor $c$ with explicit positional encoding following~\cite{xu2020positional}, which we call single-scale positional encoding (SS-PE). Surprisingly, the SS-PE in Fig.~\ref{fig:toy-noise} generates the different shapes of digits according to the 2D noise map $\epsilon$, despite the same latent vector $w$.
As mentioned previously, the constant tensor $c$ of the baseline implicitly encodes the positional information across the image. In contrast, explicit positional encoding of SS-PE captures unique positional information only for the coarsest scale (e.g., $4\times 4$). 
Eventually, zero-padding is the only component that serves as an implicit encoding at other scales, which is insufficient for providing positional information to every location. Thus, the generator has no choice but to rely on spatially varying 2D noise maps $\epsilon$ to render shapes. 

In contrast, the MS-PE provides unique positional information for every location at each scale. The generator no longer needs to rely on its implicit encoding. Fig.~\ref{fig:single-pe}(b) illustrates that explicit encoding is translation-equivariant and thus spatially unbiased. Therefore the generator successfully learns the mapping between the given positional information and the related content, thus generating consistent digits, as shown in Fig.~\ref{fig:toy}(c), even at the unseen positions. Fig.~\ref{fig:toy}(d) shows the similarity measured during successive translations. MS-PE exhibits the most consistent generation. 
The similarity of the baseline 
are lower by a large gap, suggesting that the $c$ is translation-variant and inaccurate positional encoding. 
SS-PE shows the lowest similarity since digits change rapidly according to 2D noise.

\subsection{Generation from Unseen Position and Scale}\label{sec:3.4}

MS-PE effectively learns spatially unbiased generators that generate images according to the given positional information. Therefore, a simple modification of positional encoding at inference time enables the generation of images at unseen positions and resolutions. 

Let us consider a $L$-scale generator. A single pixel shift in the $l$-th scale results in a $2^{L-l}$ pixel shift in the image space. In the opposite direction, the $k$ pixel shift in the image space corresponds to a shift of $k\times2^{l-L}$ in the $l$-th scale. Therefore we shift the positional encoding in each scale $l$ by $k\times2^{l-L}$ to generate images in $k$ pixel shifted coordinates. Following Eq.~\ref{eq:process}, $Shift_{\triangle h,~\triangle w}(I)$ is a function of
$PE_{i-\frac{\triangle h \% H}{2^{L-l}},~j-\frac{\triangle w \% W}{2^{L-l}}}^{l}$.

Sinusoidal embedding allows a decimal number of coordinates, which enables generation from continuously translated coordinates. In Fig.~\ref{fig:toy}(d), MS-PE shows consistent similarity at shifted locations by a single pixel. Meanwhile, as the constant tensor $c$ does not support shift by a decimal number, the similarity of the baseline plunge at shifts except 16 and 32. Similarly, 32 pixels shifted sample of Fig.~\ref{fig:teaser}(a) shows destructive facial features.

Moreover, explicit positional encoding can express different resolutions through a simple interpolation~\cite{xu2020positional}. This motivates multi-scale generation in Sec.~\ref{sec:4.2}, where images of multiple resolutions are generated with a model designed for a single resolution.

\section{Experiment}\label{sec:4}
We validate our method with StyleGAN2~\cite{karras2020analyzing} trained on customized color-MNIST, FFHQ~\cite{stylegan} and LSUN-Church~\cite{yu2015lsun} and show translation-equivariant generations in Fig.~\ref{fig:teaser} and Fig.~\ref{fig:toy}. We further demonstrate how a spatially unbiased generative model is beneficial for various generation tasks. Sec.~\ref{sec:4.1} demonstrates robustness in GAN inversion. Sec.~\ref{sec:4.2} presents multi-scale generation with a single model. Sec.~\ref{sec:4.3} presents generation of arbitrary sizes and aspect ratios. 
Sec.~\ref{sec:4.4} reports the quantitative results. In Sec.~\ref{sec:4.5}, we show an extension of our method to the denoising diffusion probabilistic model. See the supplementary material for details on implementation and evaluation.

\begin{figure*}[t!]
    \centering
    \includegraphics[width=0.95\linewidth]{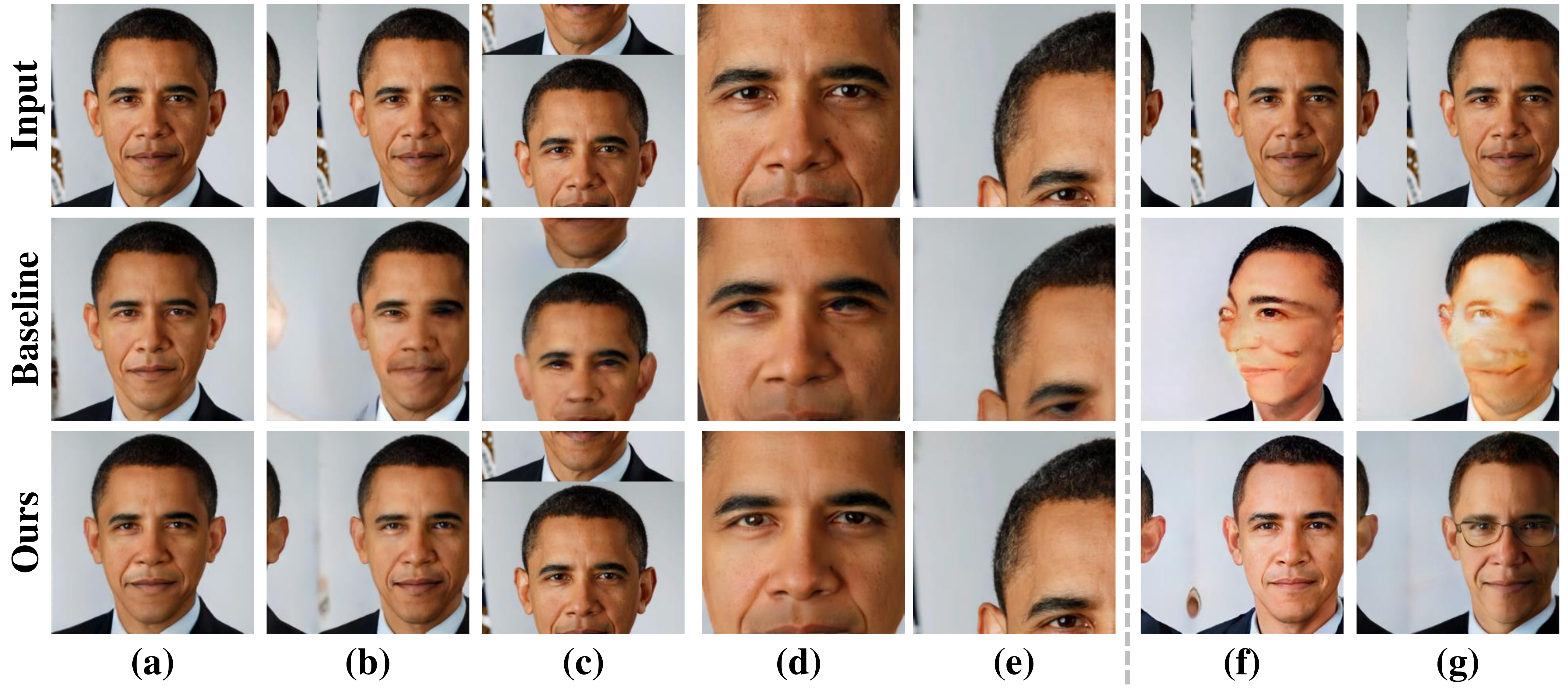}
    \vspace{-0.2em}
    \caption{\textbf{GAN Inversion.} (a) Standard position. (b) Translation of 64 pixels to the right. (c) Translation of 64 pixels to the bottom. (d) Zoom in by 2x in top-left corner. (f,g) Edition following~\cite{shen2021closedform}. Generator learned by our method show robust GAN inversion and edition at various locations and scales.}
    \label{fig:inverse}
\end{figure*}

\begin{figure*}[t!]
    \centering
    \includegraphics[width=0.95\linewidth]{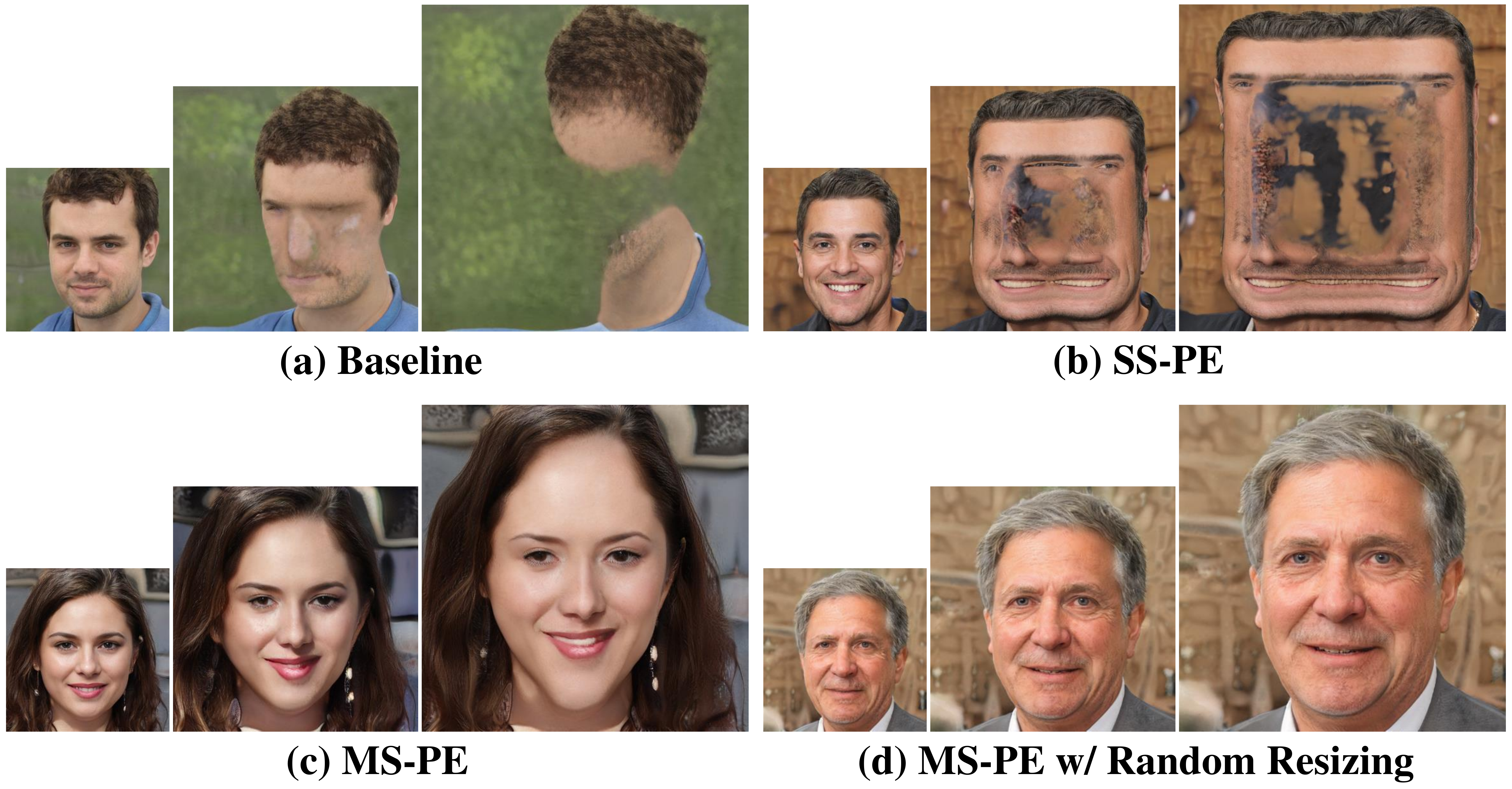}
        \vspace{-0.2em}
    \caption{\textbf{Ablation on Multi-Scale Generation}. $256^{2}$, $384^{2}$, $512^{2}$ resolution images generated with a single model. Generated by interpolating (a) constant tensor, (b) positional encoding in the coarsest scale, (c, d) positional encoding in multi-scale. Models learned by our method (c, d) generate reasonable faces in various resolutions. (b) Face structure in the boundary implies strong influence of zero-padding.}
    \label{fig:multi-scale}
\end{figure*}

\begin{figure*}[t!]
    \centering
    \includegraphics[width=0.97\linewidth]{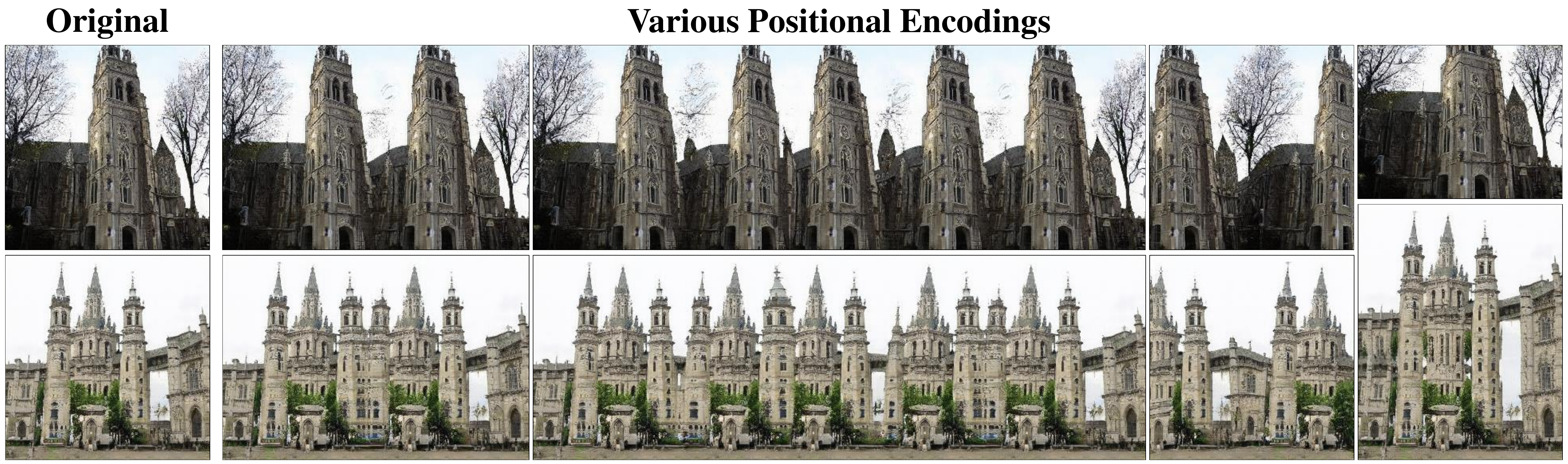}
    \vspace{-0.5em}
    \caption{\textbf{Expansion to arbitrary sizes and ratios.} Generated images expanded to arbitrary sizes and ratios while preserving their patch distribution. Towers, trees, and bridges are repeated in various locations.}
    \label{fig:expand}
\end{figure*}

\begin{figure*}[t!]
    \centering
    \includegraphics[width=0.97\linewidth]{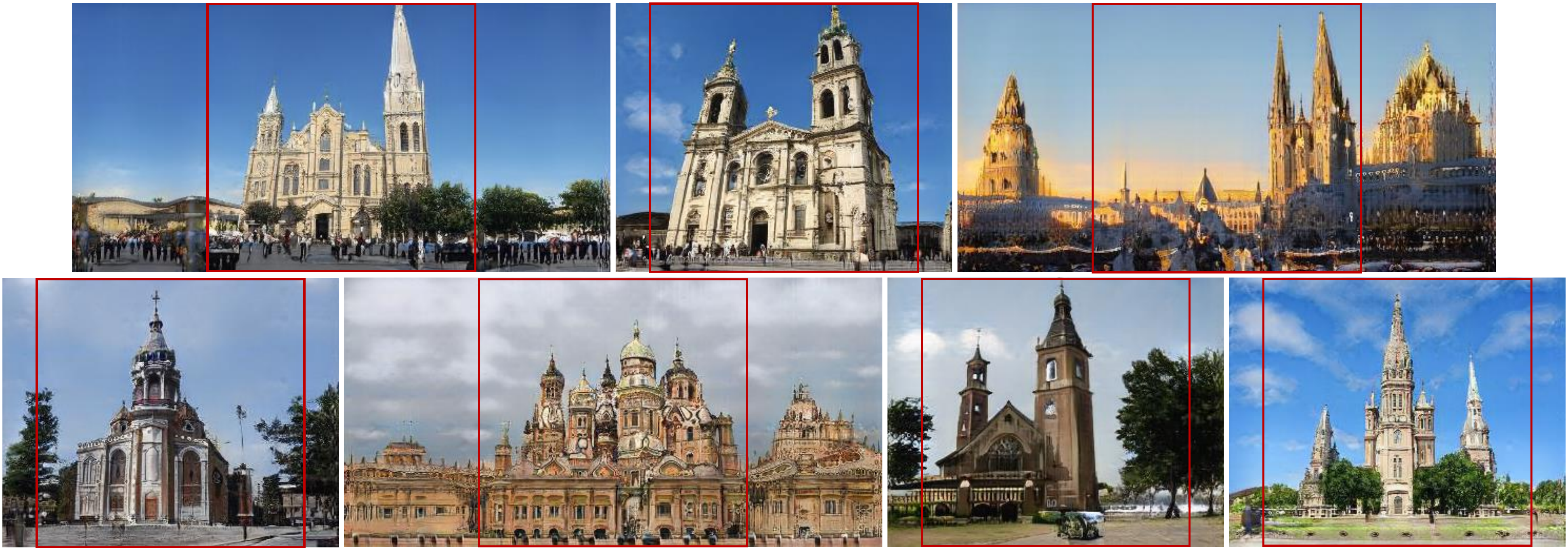}
    \vspace{-0.2em}
    \caption{\textbf{Generation beyond the boundaries.} 256x384, 256x512 resolution images generated with a model trained on 256x256 by extrapolating positional encodings beyond the boundaries. Red box denotes original area the model learned.}
    \vspace{-0.2em}
    \label{fig:extrapolate}
\end{figure*}

\subsection{GAN Inversion}\label{sec:4.1}

Recent GAN inversion literatures~\cite{abdal2019image2stylegan,huh2020transforming,gu2020image,menon2020pulse} leverage well-performed GANs on various image processing tasks. They first project~\cite{abdal2019image2stylegan,karras2020analyzing} the real image into the latent space of the GAN by optimization. Then, exploring the latent space enables various image manipulations and restorations such as style transfer~\cite{abdal2019image2stylegan}, class editing~\cite{huh2020transforming}, face editing~\cite{shen2020interpreting}, and super-resolution~\cite{menon2020pulse,gu2020image}.
However, as studied in the Im2StyleGAN~\cite{abdal2019image2stylegan}, GAN inversion is sensitive to affine transformations such as translation and resizing.

To demonstrate our robustness to transformations in GAN inversion, we optimized the StyleGAN~\cite{stylegan,karras2020analyzing} latent vector for a given image following StyleGAN2~\cite{karras2020analyzing} literature. In addition, we edited images by moving the optimized latent vector along the directions found with SeFa~\cite{shen2021closedform}. Since our method frees the generative model from spatial bias, it enables robust GAN inversion and edition, as shown in Fig.~\ref{fig:inverse}. Baseline StyleGAN fails to reconstruct unaligned face images, whereas ours reconstructs faces shifted in the vertical and horizontal directions and zoomed faces in various positions. Edited images are even more destructive than the reconstructed images. We expect these results to broaden the applicability of well-performed generative models in image-processing tasks.

\subsection{Multi-Scale Generation}\label{sec:4.2}
\vspace{-0.3em}
Fig.~\ref{fig:multi-scale} shows the qualitative comparison for multi-scale generation. As the constant tensor of the baseline leads to the generator biased to $256^2$ resolution, Fig.~\ref{fig:multi-scale}(a) shows destructive samples for the unseen scales due to the reliance on constant tensor $c$ and zero-padding. Since SS-PE lacks positional information from the previous scale as argued in Sec.~\ref{sec:3.3}, only zero-padding operates as implicit positional encoding. Therefore, the generator relies strongly on zero-padding and shows reasonable structure only at the boundary in Fig.~\ref{fig:multi-scale}(b). As shown in Fig.~\ref{fig:multi-scale}(c), our method generates reasonable global structures at various scales.
In table~\ref{table:fid_multi}, MS-PE exhibits better FID scores in all resolutions than the others by a large gap. However, generated images are awkward in detail since convolution has a scale-variant nature. Therefore, 
we train the model to generate multiple resolutions
by randomly resizing (MS-PE w/ Random Resizing) the explicit positional encoding at each iteration. Still, the discriminator learns only on a single resolution. To match the discriminator's input resolution, we resize generated image of random sizes to $256^2$. With this learning strategy, Fig.~\ref{fig:multi-scale}(d) shows consistent high-quality samples in various scales generated with a single model.

\begin{figure*}[t!]
    \centering
    \includegraphics[width=0.95\linewidth]{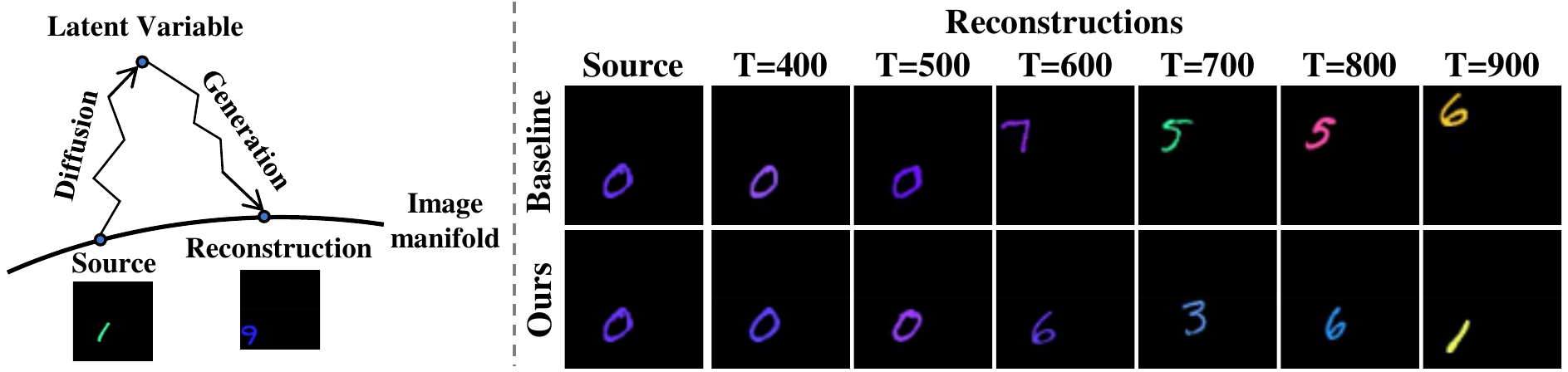}
    \caption{\textbf{DDPM Reconstructions.} From T=600, baseline DDPM generates digits in the upper-left corner as it learned spatial bias. Ours can consistently generate in given position.}
    \label{fig:ddpm}
\end{figure*}

\subsection{Generation of arbitrary sizes and aspect ratios}\label{sec:4.3}

From Sec.~\ref{sec:3.3}, the generator learns a mapping between explicit positional encoding and the corresponding content. This property motivates generation from expanded coordinate grids of arbitrary sizes and aspect ratios. 
By repeating the desired part of the positional encoding, images of arbitrary sizes and aspect ratios can be generated while preserving their patch distribution. Fig.~\ref{fig:expand} shows samples of various resolutions generated with a single model trained at a resolution of $256^2$. As can be observed, towers, trees, and bridges are rendered in various locations and orders. Expanding generated images is studied in SinGAN~\cite{shaham2019singan}, which train an unconditional model on a single image. However, we used a model trained on a large dataset. 

Our method not only expands images but also enables generation beyond the image boundary. Since our method learns a spatially unbiased generator, the generator can also generate contents beyond the learned position. We trained model on the coordinate grid $[0,H^l]\times [0,W^l]$ for each scale $l$, and we can generate from the coordinate grid, for example, $[0,H^l]\times [-0.5W^l,1.5W^l]$ at each scale. Fig.~\ref{fig:extrapolate} shows realistic samples even beyond the boundary. Extra objects such as trees, humans, buildings are generated. A similar experiment on extrapolation was presented in~\cite{liu2018intriguing,lin2019coco}, however, our work differs in that we generated a full-resolution image with an unconditional model.

\subsection{Quantitative Results}\label{sec:4.4}

We evaluated our method with Fréchet Inception Distance (FID)~\cite{heusel2017gans} as well as Precision and Recall~\cite{sajjadi2018assessing}. We compare our MS-PE with baseline and SS-PE, those described in Sec.~\ref{sec:3.1}. Our method shows better performance in FFHQ~\cite{stylegan}, suggesting that the model learned rich representation from spatial bias in the dataset while not relying on it. Recall increased rapidly, suggesting that unbiased learning led to improved diversity. Our method maintains performance in LSUN-Church~\cite{yu2015lsun}, while enabling diverse generations as shown in Sec.\ref{sec:4.3}.

\subsection{Additional Experiment on Denoising Diffusion Probabilistic Models}\label{sec:4.5}

Denoising Diffusion Probabilistic Models (DDPM)~\cite{ho2020denoising,sohl2015deep} is a recently introduced family of generative models which have shown powerful generation performance in various domains~\cite{ho2020denoising,nichol2021improved,chen2020wavegrad,kong2020diffwave}. It learns the reverse of the \textit{diffusion process} where 2D noise maps are gradually added to the image. Since each step of the generation process (1000 steps in total) is stochastic, highly diverse samples are generated from the same latent variable. See supplementary materials for details on DDPM.

Ho~\etal~\cite{ho2020denoising} presented stochastic reconstruction, where images are encoded into an intermediate time step and then decoded by a learned generation process. Both encoding and decoding processes are stochastic. Therefore, as the image is encoded into a larger time step, more diverse samples are generated. We learn the DDPM on color-MNIST biased to the upper-left corner, then provide inputs biased to the bottom-left corner. In Fig.~\ref{fig:ddpm}, when reconstructed from time step over 500, baseline generates digits in the upper-left corner. This indicates that the baseline DDPM is spatially biased. However, the model learned by our method reconstructs digits at the bottom-left, which is an unseen position. This example indicates that MS-PE can also be applied to learn spatially unbiased DDPM.

\begin{table}[]
\centering
\begin{tabular}{l|ccc}
\Xhline{1pt}\\[-0.95em]
    Config      & @256 & @384 & @512 \\ \hline\hline
Baseline   &  7.21    &  43.11    &  154.40    \\
SS-PE &   7.37   &   94.44   &   187.78   \\
MS-PE (Ours)      &   \textbf{6.75}   &   \textbf{16.03}   &   \textbf{30.41}   \\
 \Xhline{1pt}
\end{tabular}
\caption{\textbf{Ablation on multi-scale generation.} FID scores with $256^2$, $384^2$, $512^2$ images generated by $256^2$ models. Our method shows superior performances.}
\label{table:fid_multi}
\end{table}

\begin{table}[]
\centering
\begin{tabular}{l|ccc|ccc}
\Xhline{1pt}
              & \multicolumn{3}{c|}{FFHQ} & \multicolumn{3}{c}{LSUN-Church} \\ \hline
Config & FID$\downarrow$  & P$\uparrow$ & R$\uparrow$ & FID$\downarrow$    & P$\uparrow$   & R$\uparrow$   \\ \hline
Baseline      &   7.21   &     64.07      &   35.20     &   \textbf{4.70}     &     59.74        &     \textbf{39.74}     \\
SS-PE    & 7.37     &    \textbf{64.19}       &    36.97    &   5.01     &     \textbf{60.53}        &     36.43     \\
MS-PE        &   \textbf{6.75}   &     62.97      &   \textbf{42.27}     &    6.42    &       55.91     &    34.20      \\ \Xhline{1pt}
\end{tabular}
\caption{\textbf{Quantitative comparison.} FID, Precision \& Recall (\%) calculated with $256^2$ images.}
\label{table:fid}
\vspace{-1em}
\end{table}

\section{Conclusion}
We introduced a simple method for learning spatially unbiased generative models. We demonstrated that generators rely on translation-variant implicit positional encoding to generate spatial content. This caused spatially biased generator, which show degraded performance at unseen position and scale. By injecting explicit positional encoding in each scale, we learn generators that render images according to the given positional information. Concurrent work~\cite{karras2021alias} presented FFHQ-U, unaligned version of existing FFHQ~\cite{stylegan} dataset. We are looking forward to training on such unaligned dataset for future work.


\setcounter{section}{0}
\renewcommand\thesection{\Alph{section}}
\setcounter{table}{0}
\renewcommand{\thetable}{\Alph{table}}
\setcounter{figure}{0}
\renewcommand{\thefigure}{\Alph{figure}}
\setcounter{equation}{0}
\renewcommand{\theequation}{\Alph{equation}}

\section{Additional Figures}
Here, we show additional qualitative results.

\begin{figure}[t!]
    \centering
    \includegraphics[width=1.0\linewidth]{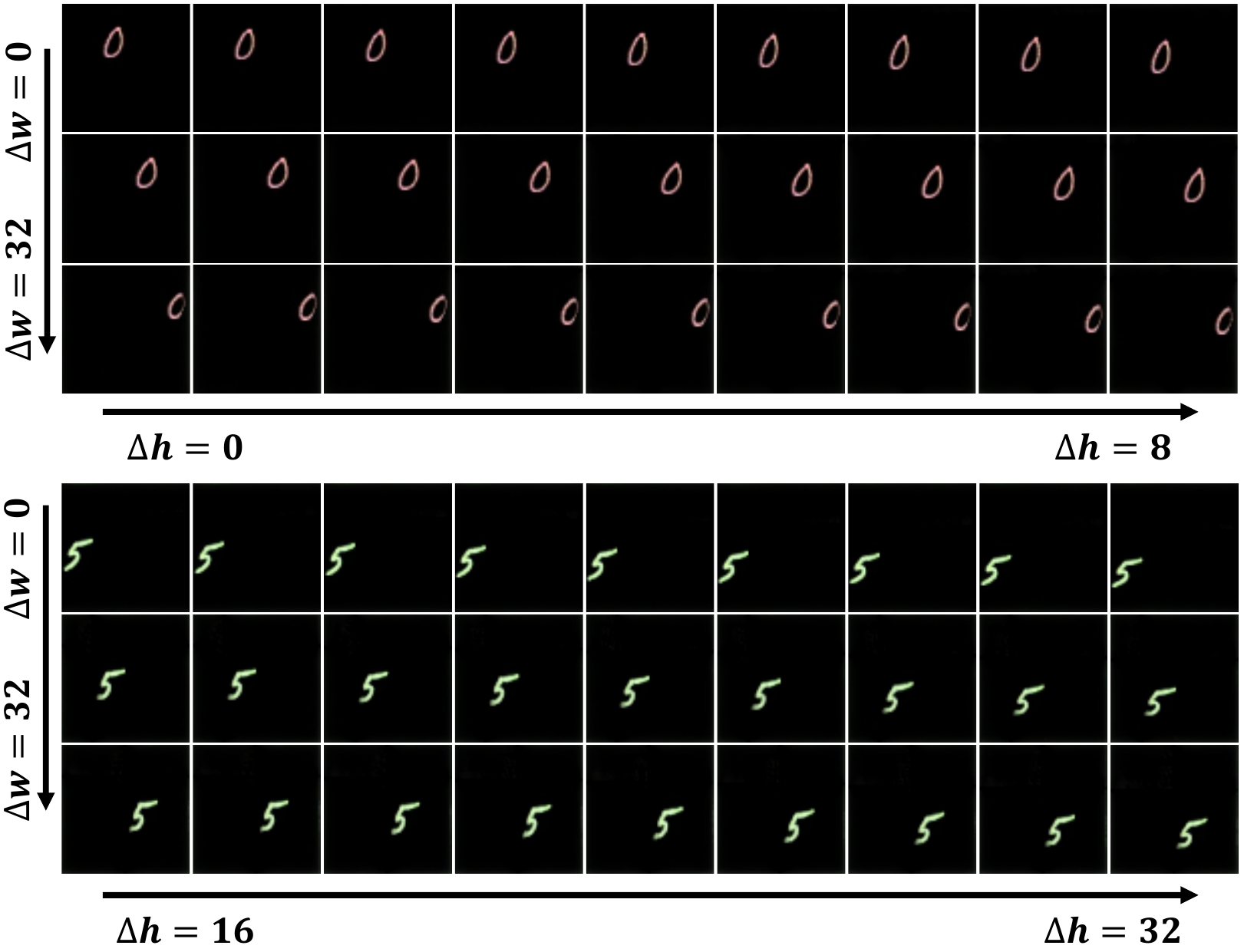}
    \caption{\textbf{Generation at successively shifted locations.} We generated digits at locations shifted vertically by $\triangle h$ and horizontally by $\triangle w$, sharing the same inputs. Digits show consistent shapes.}
    \label{fig:finefine}
\end{figure}

\textbf{Successive translations} In the main text, we discussed that sinusoidal embedding enables non-integer coordinates. Thus MS-PE generates consistent shapes at locations successively shifted by a pixel.  Fig.~\ref{fig:finefine} show digits generated at locations shifted vertically and horizontally. 

\begin{figure}[t!]
    \centering
    \includegraphics[width=1.0\linewidth]{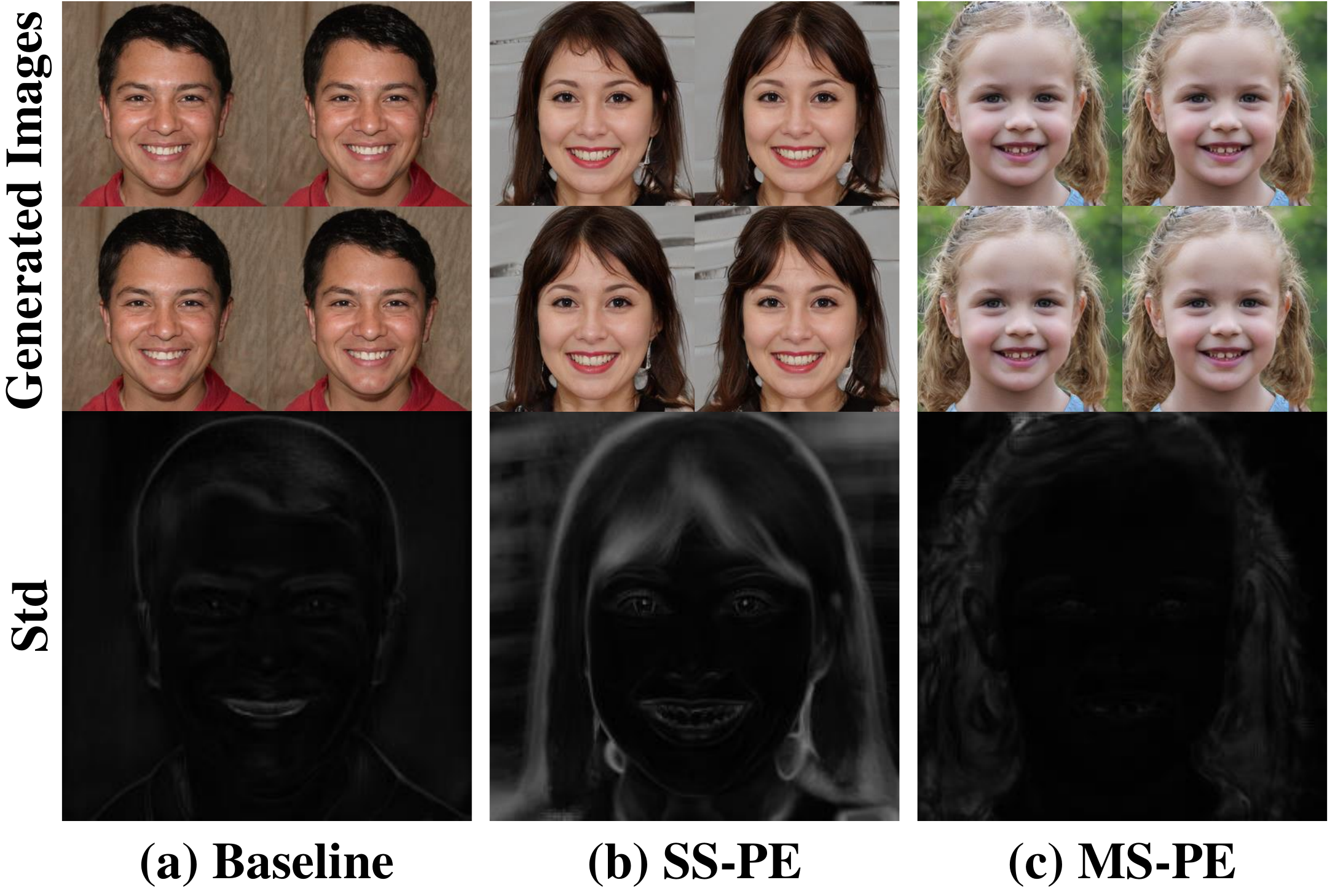}
    \caption{\textbf{Effect of 2D noise input.} Standard deviation of each pixel over 100 2D noise map instances. SS-PE generates inconsistent hairstyles and larger standard deviations compared to the baseline and MS-PE.}
    \label{fig:ffhq-noise}
\end{figure}

\textbf{Effect of 2D noise} SS-PE rely on 2D noise input of StyleGAN~\cite{stylegan,karras2020analyzing}, as shown in Fig.4 of the main text. Fig.~\ref{fig:ffhq-noise} exhibits additional results with models trained on FFHQ~\cite{stylegan}. In Fig.~\ref{fig:ffhq-noise}(b), the hairstyle is modified by a 2D noise map and shows the largest standard deviation.

\begin{figure*}[t!]
    \centering
    \includegraphics[width=1.0\linewidth]{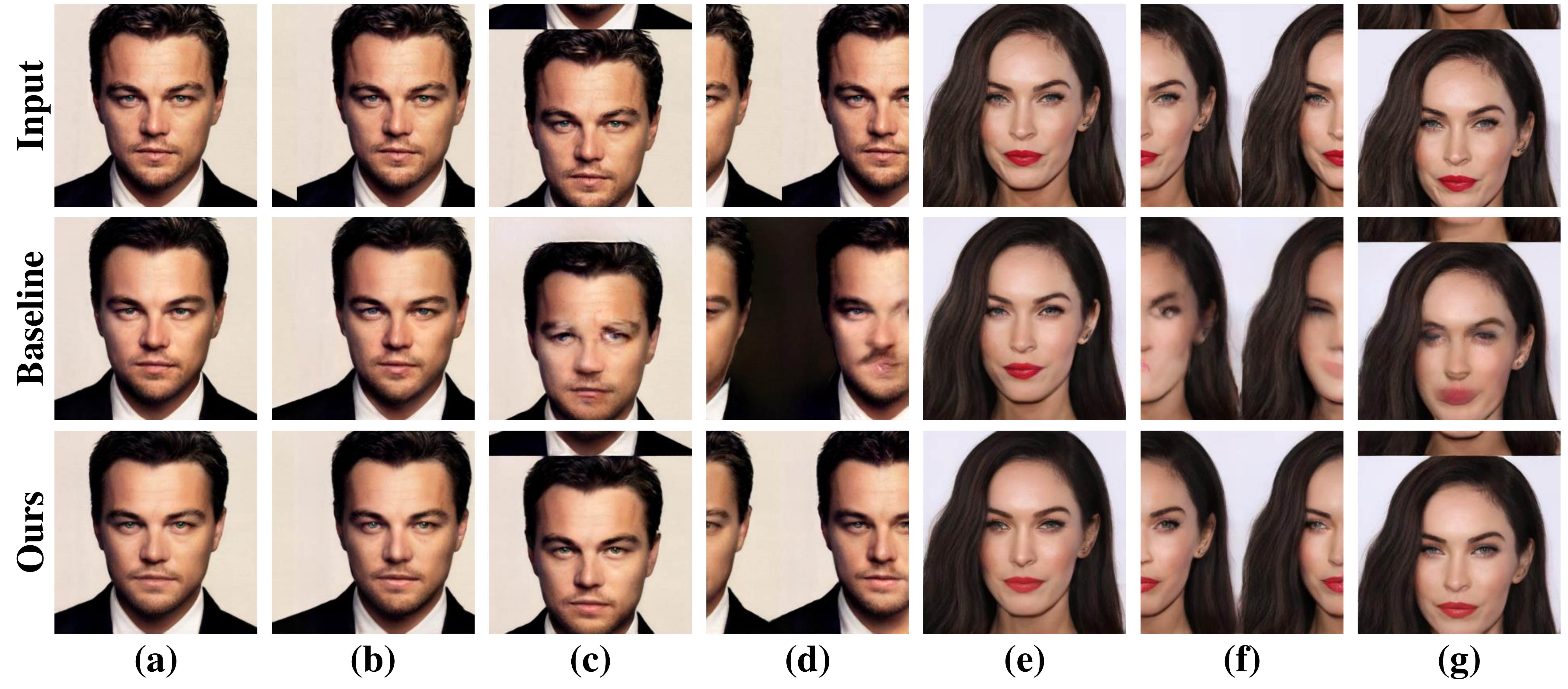}
    \caption{\textbf{GAN Inversion.} (a) Standard position. (b) Translation of 32 pixels to the right. (c) Translation of 32 pixels to the bottom. (d) Translation of 128 pixels to the right. (e) Standard position. (f) Translation of 128 pixels to the right. (g) Translation of 32 pixels to the bottom. We use circular shift (roll) for translations.}
    \label{fig:inversesupp}
\end{figure*}

\textbf{GAN Inversion}
Our method facilitates robust GAN inversion. Additional GAN inversion results are presented in Fig.~\ref{fig:inversesupp}. 

\begin{figure*}[t!]
    \centering
    \includegraphics[width=0.9\linewidth]{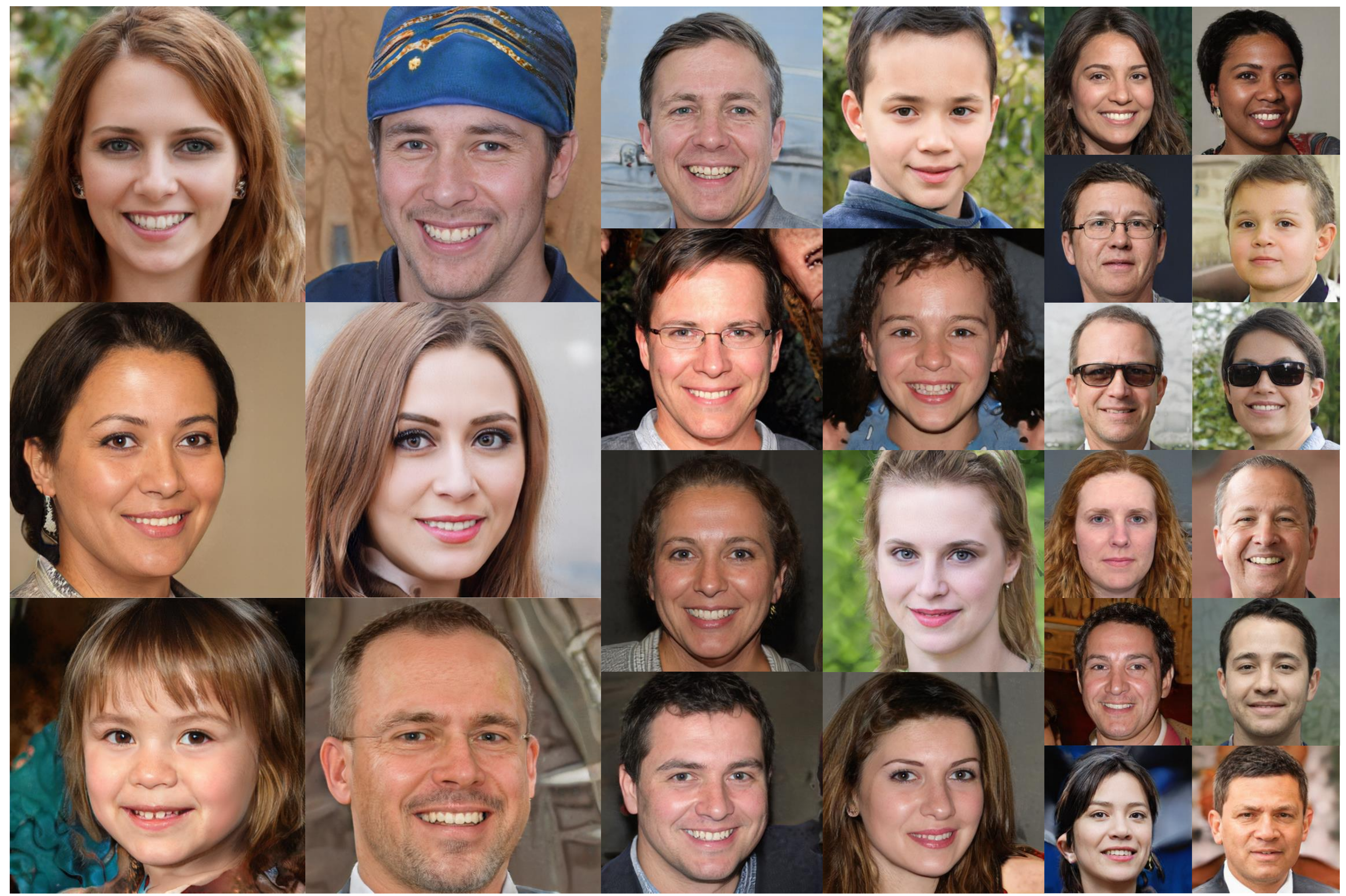}
    \caption{\textbf{Multi-scale generation.} $256^{2}$, $384^{2}$, $512^{2}$ resolution images generated with a single model designed for $256^{2}$ . Discriminator has seen images at $256^{2}$ resolution only.}
    \label{fig:multisupp}
\end{figure*}

\textbf{Multi-scale generation} MS-PE is effective in multi-scale generation with a single model. To further improve visual quality, we randomly resized (MS-PE w/ Random Resizing) the explicit positional encoding at each training iteration. Additional samples with ``MS-PE w/ Random Resizing'' are presented in Fig.~\ref{fig:multisupp}.

\begin{figure*}[t!]
    \centering
    \includegraphics[width=0.9\linewidth]{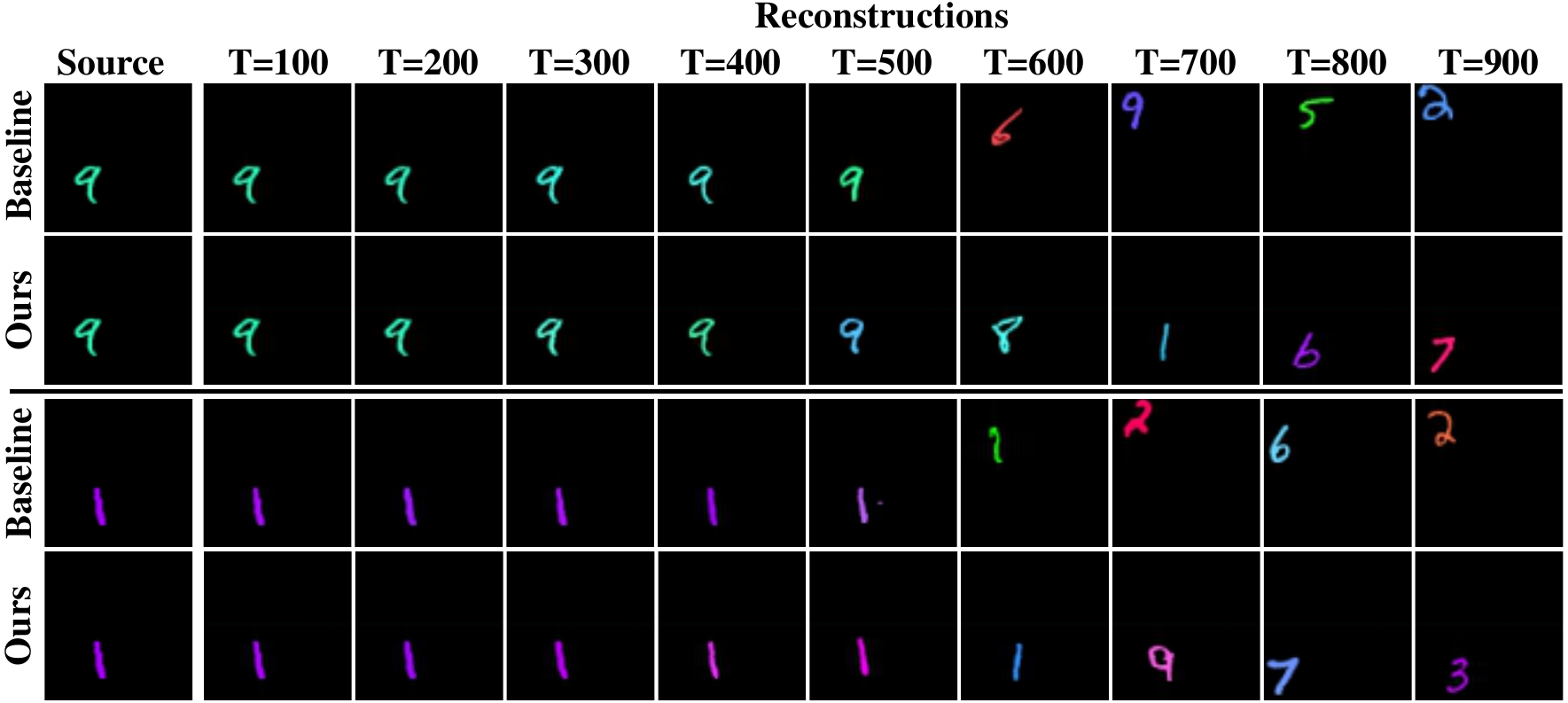}
    \caption{\textbf{DDPM Reconstruction.} Given source images are encoded to various time steps (T=100 to 900) then decoded with a learned reverse process. Our method preserves location of the given digits.}
    \label{fig:ddpmsupp}
\end{figure*}

\textbf{DDPM Reconstruction}
In the main text, we demonstrated the appliance of our method to denoising diffusion probabilistic models~\cite{ho2020denoising,sohl2015deep}.  Fig.~\ref{fig:ddpmsupp} shows additional reconstruction results.

\section{Detailed Introduction to DDPM}
Here, we provide an additional description on Denoising Diffusion Probabilistic Models (DDPM)~\cite{ho2020denoising,sohl2015deep}. 
DDPM consists of two processes: fixed \textit{diffusion process} and learned \textit{reverse process}. The diffusion process is a sequence that adds Gaussian noise to an image $x_0$ with a fixed variance schedule $\beta_1,...,\beta_T$. Latent variables $x_1,...,x_T$ are sampled from the diffusion process:
\begin{equation}\label{eq:forward}
q(x_{t}|x_{t-1}):=N(x_{t};\sqrt{1-\beta _{t}}x_{t-1},\beta _{t}\mathbf{I}).
\end{equation}
To generate images from random Gaussian noise, DDPM learns the reverse of the diffusion process, which is also a sequence of Gaussian translation:
\begin{equation}\label{eq:reverse}
p_{\theta}(x_{t-1}|x_{t})=N(x_{t-1};\mu_{\theta}(x_{t},t),{\sigma}_t^2\mathbf{I}).
\end{equation}
Here, variance is a fixed constant~\cite{ho2020denoising} and the mean is learned with a neural network $\theta$. Here, $\theta$ is a fully-convolutional U-Net~\cite{ronneberger2015u} with input and output of the same dimensionality. As diffusion (Eq.~\ref{eq:forward}) and reverse (Eq.~\ref{eq:reverse}) process are both Gaussian translations, they are stochastic.
Sec.4.5 presents stochastic reconstruction where encoding (diffusion) process and decoding (reverse) process are both stochastic. 

\section{Implementation Details}

\subsection{Dataset}
\textbf{Color-MNIST} is a spatially biased dataset we customized for toy experiments presented in Fig.2, Fig.4, Fig.9 of the main text. The dataset consists of 60,000 images at $64\times 64$ resolution. Digits range from 0 to 9, and their numbers are uniform. Digits are located in the $32\times 32$ patch at the upper-left corner.

\textbf{Flickr Faces HQ~\cite{stylegan}} consists of 70,000 high-quality face images crawled from Flickr. The images are carefully aligned~\cite{kazemi2014one}, thus exhibits strong location bias. 

\textbf{LSUN Church~\cite{yu2015lsun}} consists of 126,227 images of churches. Center cropped $256\times 256$ images are used for training models.

\subsection{Architecture}
\textbf{StyleGAN2} For experiments except Sec.4.5 of the main text, we used StyleGAN2~\cite{karras2020analyzing} trained at $256\times 256$ resolution and inherited most of the architecture details of StyleGAN2. The details include weight demodulation, bilinear up/down sampling~\cite{zhang2019making}, noise injection, skip/residual connection in generator/discriminator, equalized learning rate, leaky ReLU activation with slope 0.2, and minibatch standard deviation at the discriminator. These setups correspond to the ``baseline'' in the main text. We replaced the constant tensor with 2D sinusoidal positional embedding~\cite{vaswani2017attention}. We also added scale-specific 2D sinusoidal positional embedding at each scale, as described in Eq.3 of the main text. Therefore seven positional encodings in total (from $4\times 4$ to $256\times 256$). 

\textbf{DDPM} In Sec.4.5 of the main text, we inherit the architecture details of DDPM~\cite{ho2020denoising}, including U-Net~\cite{ronneberger2015u} architecture, group normalization~\cite{wu2018group}, self-attention blocks, linear $\beta _t$ schedule, and sinusoidal embedding to indicate time step. Our $64\times 64$ model use four feature map resolutions ( $64\times 64$ to $8\times 8$) and self-attention blocks at $8\times 8$ resolution. We added 2D sinusoidal positional encoding after residual blocks at downscaling layers and after upsampling layers.

\subsection{Training}
\textbf{StyleGAN2} For every configuration, we trained models for 6.4M images with batch size 32. We used non-saturating loss~\cite{goodfellow2014generative} with $R_1$ regularization ~\cite{mescheder2018training}. We used only random horizontal flip for data augmentation. We used ADAM~\cite{kingma2014adam} optimizer with $\beta_1=0, \beta_2=0.99$. For MS-PE with Random Resizing in Sec.4.2, we randomly selected resolution from $\{256^2, 320^2, 384^2, 448^2, 512^2\}$ with uniform probabilities at each iteration. 

\textbf{DDPM} We trained DDPM on our color-MNIST for 9.6M images with batch size 16. We did not use any data augmentations. We used ADAM~\cite{kingma2014adam} optimizer with $\beta_1=0, \beta_2=0.99$.

\subsection{Evaluation}
\begin{figure}[t!]
    \centering
    \includegraphics[width=0.8\linewidth]{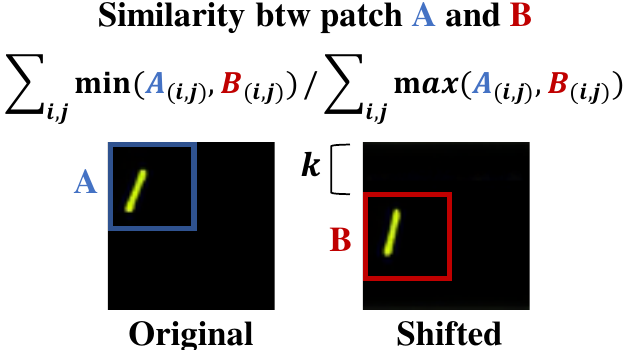}
    \caption{\textbf{Measuring similarity of patches.} It is a continuous relaxation of mIoU. $i,j$ are spatial indexes where $0\leq i,j\leq31$.}
    \label{fig:similarity}
\end{figure}

\textbf{Similarity metric} In Fig.2(d) of the main text, we measured the similarity of digits during successive translations. To compare the original digit with the shifted digit, we crop $32\times 32$ patch A and B, as shown in Fig~\ref{fig:similarity}. We then convert them to grayscale and measure similarity as follows:
\begin{align}\label{eq:similarity}
sim=\sum_{i,j}min(A_{i,j},B_{i,j})/\sum_{i,j}max(A_{i,j},B_{i,j}),
\end{align}
where $i,j$ are spatial indexes. This metric is a continuous relaxation of mean Intersection over Union (mIoU).

\begin{figure}[t!]
    \centering
    \includegraphics[width=0.7\linewidth]{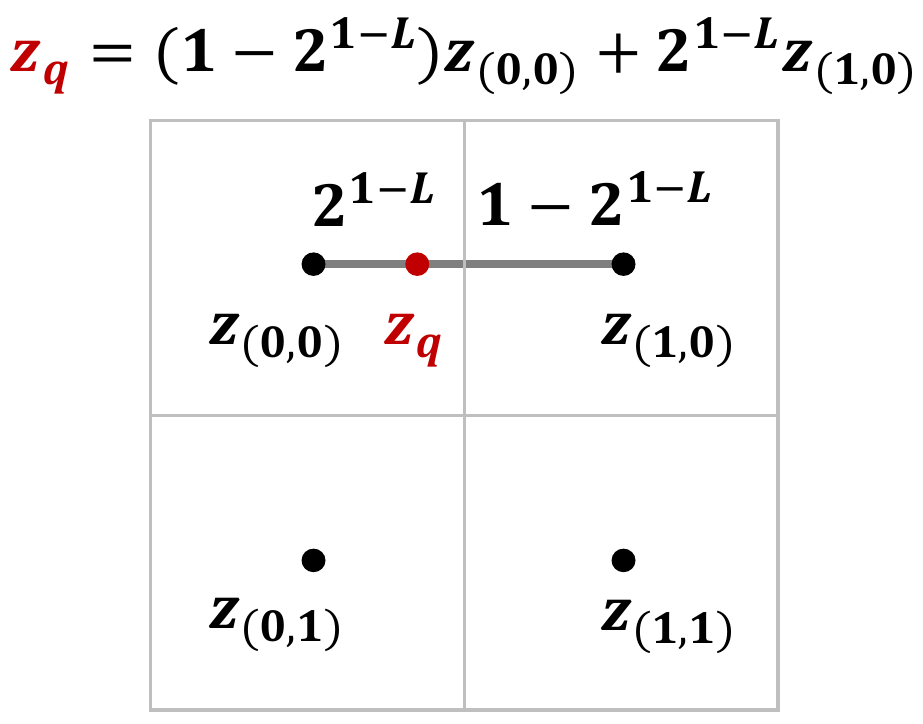}
    \caption{\textbf{Interpolation at the constant tensor.} To implement non-integer shift in the constant tensor of baseline StyleGAN, we interpolated features at nearest integer coordinates. $Z$s denote features at each coordinate. We implement a one pixel shift of image by replacing $z_{(0,0)}$ with $z_{q}$.}
    \label{fig:interpolation}
\end{figure}

\textbf{Shift in the constant tensor.} Fig.2(d) of the main text presents successive translations by a pixel. As described in Sec.3.4, a single-pixel shift in the image space corresponds to a $2^{1-L}$ shift in the constant tensor of the baseline StyleGAN ($L$-scale). To implement a non-integer shift in the constant tensor, we interpolated features of nearest integer coordinates, as shown in Fig.~\ref{fig:interpolation}.

\textbf{Fréchet Inception Distance (FID)}~\cite{heusel2017gans} We calculated FID scores on 50,000 real images and 50,000 generated images using code\footnote{\url{https://github.com/mseitzer/pytorch-fid}} of the PyTorch framework.

{\small
\bibliographystyle{ieee_fullname}
\bibliography{egbib}
}

\end{document}